# Integrated Design-Fabrication and Control of a Bio-Inspired Multi-Material Soft Robotic Hand.

Samuel Alves, Mihail Babcinschi, Afonso Silva, Diogo Neto, Diogo Fonseca, and Pedro Neto[*]

*University of Coimbra, CEMMPRE, ARISE, Department of Mechanical Engineering, 3030-788, Coimbra, Portugal*
*Address correspondence to: pedro.neto@dem.uc.pt

**Abstract**

Machines that mimic humans have inspired scientists for centuries. Bio-inspired soft robotic hands are a good example of such an endeavor, featuring intrinsic material compliance and continuous motion to deal with uncertainty and adapt to unstructured environments. Recent research led to impactful achievements in functional designs, modeling, fabrication, and control of soft robots. Nevertheless, the full realization of life-like movements is still challenging to achieve, often based on trial-and-error considerations from design to fabrication, consuming time and resources. In this study, a soft robotic hand is proposed, composed of soft actuator cores and an exoskeleton, featuring a multi-material design aided by finite element analysis (FEA) to define the hand geometry and promote finger's bendability. The actuators are fabricated using molding and the exoskeleton is 3D-printed in a single step. An ON-OFF controller keeps the set fingers' inner pressures related to specific bending angles, even in the presence of leaks. The FEA numerical results were validated by experimental tests, as well as the ability of the hand to grasp objects with different shapes, weights and sizes. This integrated solution will make soft robotic hands more available to people, at a reduced cost, avoiding the time-consuming design-fabrication trial-and-error processes.

1. **Introduction**

Robot actuators made of rigid materials are precise and controllable. However, their compliance to accommodate uncertainty is limited and its high stiffness can cause injuries to humans at shared workspaces or damage equipment in case of unexpected collisions. Inspired by nature [1], soft robotics overcome the issues posed by traditional robot actuators, promoting a smooth and safe interaction with the surrounding environment due to its intrinsic compliance and flexibility, which is especially relevant when operating in unstructured environments [2,3]. Rigid (stiff) materials can be replaced by elastomers, promoting the robot's continuous motion and ability to adapt itself to the environment. Silicone-bodied pneumatic robot's kinematics is highly affected by the shape and material of the actuator, where motion is generated by a change of pressure in the actuator's chambers. McKibben's artificial muscle is a representative example where the radial expansion of the pressurized soft structure creates linear motion [4,5]. Soft actuators can be tethered or untethered [6–8], driven by fluid pressure and displacement [9], heat [10], magnetic fields [11,12], combustion [13], or even light [14]. While soft robots represent a new paradigm in robotics, their design, modeling, fabrication and control are scientifically and technologically challenging [15]. Frequently, soft robots' design is based on trial-and-error experiments involving the fabrication of multiple soft robot prototypes, following a cycle of testing, re-design and fabrication of an updated prototype. A significant part of such design work can be done offline, using FEA to support the design process, saving time and resources.

Multi-material pneumatic soft actuators take advantage of integrating different materials with distinct stiffness values. Accordingly, they demonstrated effective compliant behavior and dexterity, providing translation and rotation movement to bend in any direction [16], as well as adaptability to grasp different objects [17]. Antagonistic pneumatic actuators with parallel chambers enable variable stiffness while keeping the design simple but challenging to fabricate [18]. With the improvement of computational power, the numerical simulation has become a useful method in the design of soft robots. However, the FEA presents three non-linearities: (i) hyper-elastic behavior of the soft materials; (ii) finite rotation and large strain of the actuators; (iii) frictional contact between different components of the assembly. Thus, the non-linear FEA involves a high computation cost and experiences difficulties in dealing with complex non-linear contact boundary conditions, which often lead to convergence problems. The large deformations and distortions concentrated in specific areas, namely in thin elements, are challenging to FEA. Moreover, the required mechanical characterization of the soft materials is challenging due to the large strains achieved under different load paths and the time-dependent deformation behavior. Recent studies aim to improve the level of accuracy obtained in modelling of both static and dynamic behavior of soft materials [19–22]. Design optimization constraints related to stress, mass, volume and fabrication process have a key role in the simulation loop of hyperelastic multi-materials [23]. Despite this challenging context, some FEA physics-based simulators have been developed to support the design and optimization of soft robots [24–26].

Soft robots can be fabricated by using multiple materials [27] and using different manufacturing processes, ranging from silicone molding to 3D printing [28–30]. Using sequential molding, the internal chamber of soft actuators can be limited to simple geometries, taking a relatively long time to fabricate. On the other hand, 3D printing methods bring significant benefits in design and fabrication, making it easy to introduce complex geometries within soft robots, accelerating/automating the fabrication process, and reducing its cost [31–33]. The fabrication in a single step, as 3D printing, is highly desirable, allowing the introduction of sensing and control elements within the robot, promoting innovation in multiple application domains [34–36].

Soft actuator control is still far away from the motion control observed in biological systems [37,38]. Recent studies rely on logic loops with fixed and varying rate quasi-static controllers, speeding up or delaying inflation/deflation [39–42]. An actuator control system can receive feedback from the actuator's internal pressure or soft strain/displacement stretchable sensors embedded in the actuator [43,44].

Here, we present a bio-inspired soft robotic hand similar to the human hand, with five fingers pneumatically actuated. The multi-material soft actuators are designed and fabricated at a reduced cost and time effort, using standard fabrication processes such as molding and single-step 3D printing. The ON-OFF controller, while simple, keeps the set fingers' bending angles stable, even in the presence of leaks. The robotic hand demonstrated dexterity and capability to grasp objects with different shapes, weights and sizes.

## 2. Materials and Methods
### 2.1 Operating principles and design
The proposed soft robotic hand was developed by taking advantage of multiple materials, exploring the capabilities of actual 3D printing techniques and the advance of numerical modeling. Our goal is to fabricate a functional and low-cost soft robotic hand that is identical in shape and size to the human hand, Fig. 1. It is composed of a single exoskeleton and five pneumatic actuator cores, one per finger. Previous studies have demonstrated that three-fingered hands are effective to achieve arbitrary manipulation of objects and stable grasping [45,46]. Thus, the proposed hand has only three controllable elements, i.e., the thumb and index finger are controlled independently, while the middle, ring and little fingers are controlled simultaneously.

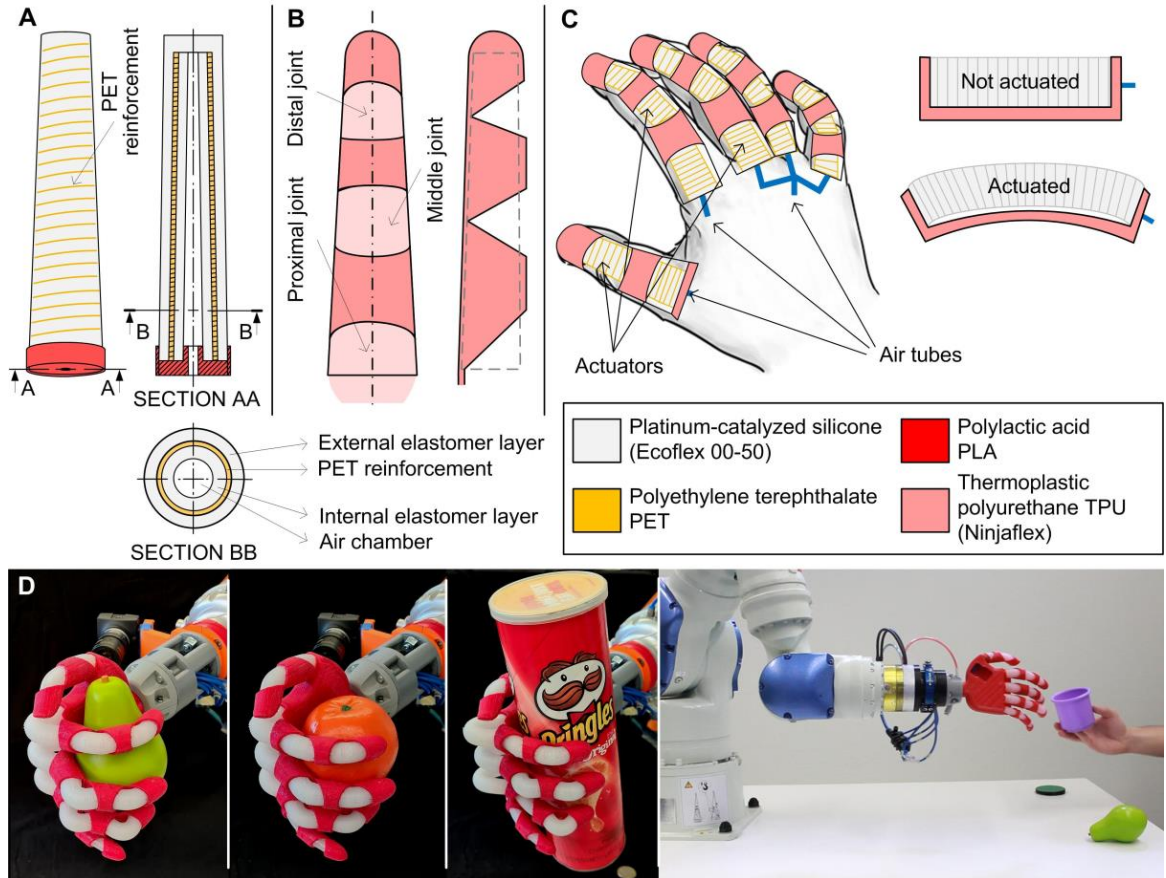

**Fig. 1**. Design principles of the soft robotic hand. (A) Multi-material finger's soft actuator composed of two silicone layers and internal PET reinforcement. (B) Exoskeleton geometry of a single finger designed to bend in 3 joints (distal, middle and proximal). (C) Operation principle where the pneumatic actuator inside a stiffer exoskeleton shell promotes the bending of the finger. (D) Attached to a robot manipulator, the soft robotic hand is capable of grasping and manipulating objects of various shapes, weights, and sizes.

Finger's bending is achieved by pressurizing the actuator, which presents a single longitudinal-cylindrical chamber. The soft actuator, made of stretchable platinum-catalyzed silicone, was reinforced by a long thread of polyethylene terephthalate (PET) wrapped around the circumferential direction, Fig. 1A. The PET reinforcement avoids the barreling effect (radial expansion) of the actuator when the airflow enters the chamber, promoting axial elongation. The exoskeleton was made of flexible thermoplastic polyurethane (TPU) and presents a small bending stiffness in the finger bending direction, which is dictated by the geometry of the fingers, Fig. 1B. Thus, the axial elongation of the actuator induces the bending motion of the exoskeleton geometry, Fig. 1C. The hand has a mass of about 100 grams, excluding the control elements and the robot tool changer, and can successfully grasp objects with different shapes, weights and sizes, Fig. 1D. When the actuator is subject to high deformations that cannot be accommodated by the exoskeleton, it can be pushed out of the exoskeleton on the proximal joint area, Fig. 1D (grasping Pringles tube). It is a compensation mechanism to "absorb" the actuator excessive elongation, preventing its rupture. Its versatility and applicability have been demonstrated by the successful integration into a robot manipulator to grasp and manipulate different objects from the Yale-CMU-Berkeley (YCB) dataset (Movie S1, Supplementary Materials). Nevertheless, the grasping performance depends on factors such as the grasping force, contact geometry and the static friction.

**2.2 Numerical modeling and simulation**
The performance of the soft robotic hand is influenced by several key design parameters, namely the geometry of each component, the materials, and the applied pressure on the actuator's chamber. Both the shape and size of the hand were predetermined to be similar to the human hand, while the wall thickness of the actuator and exoskeleton were constrained by the fabrication processes. To ensure structural integrity, a minimum wall thickness of 2 mm was defined for both the actuator and the exoskeleton. The materials were selected considering their rubber-like elastic behavior, fabrication constraints, availability and cost. The pressure applied on the actuator is a key parameter evaluated by numerical simulation. It directly affects the bending magnitude of the fingers and gripping force. However, the FEA of soft materials using hyper-elastic models is challenging to apply due to the highly non-linear stress-strain response and the large deformations involved. Moreover, the complex frictional contact between the actuators and the exoskeleton can lead to severe convergence issues while being computationally costly. The numerical simulation was adopted to quantify the relationship between the applied pressure on the actuator and the bending angle of the finger. Since all fingers have identical geometry and working conditions, only the index finger comprising a soft actuator and an exoskeleton is studied. Each component, actuator and exoskeleton, is studied independently by FEA. Previous studies demonstrated that the Mooney-Rivlin (MR) model presents a better performance for modeling elastomers with higher shore hardness, while the Ogden model is more suitable for modeling softer silicones [47–49]. Thus, the MR model was used to describe the exoskeleton material behavior, while the Ogden model was applied to define the actuator material behavior.

The barreling effect of the actuator was prevented through PET reinforcement applied in the form of a cylindrical winding. Nevertheless, the modeling of this reinforcement was simplified in the numerical simulation to avoid the modeling of hundreds of turns of thread, Fig. 2A. Accordingly, the PET reinforcement was modeled by applying 44 equally spaced rings of rectangular cross-section with 0.25 mm2 of area, Fig. 2A. The numerical results demonstrated that the PET reinforcement promotes the actuator elongation as the actuator's internal pressure increases, restricting the radial displacement, Fig. 2B. To consider the variability in the mechanical behavior of the materials, the material of the actuator was modeled using three different sets of Ogden parameters (Table S1, Supplementary Materials). Ogden set1 refers to the material with higher stiffness, whereas Ogden set3 refers to the material with lower stiffness. Using the material parameters from the Ogden set3, the linear elongation of the actuator created by the internal pressure rise is shown in Fig. 2C. The increase of the internal pressure leads to an increase of the actuator length while the PET reinforcement effectively restricted the radial displacement of the actuator. The slight curvature visible in the experimental elongation is caused by unevenness of the PET reinforcement, which was wound by hand. This does not seem to significantly impact performance according to the numerical results and could be minimized by employing an automatic winding method.

The numerical analysis of the exoskeleton evaluates its stiffness under bending. The exoskeleton was fixed at the end, near the proximal joint, and a vertical load was applied at the distal joint, Fig. 2D. The mass of the exoskeleton was taken into account in the numerical simulation as the self-weight causes deflection. Three different sets of MR parameters were considered for the material of the exoskeleton (Table S2, Supplementary Materials).

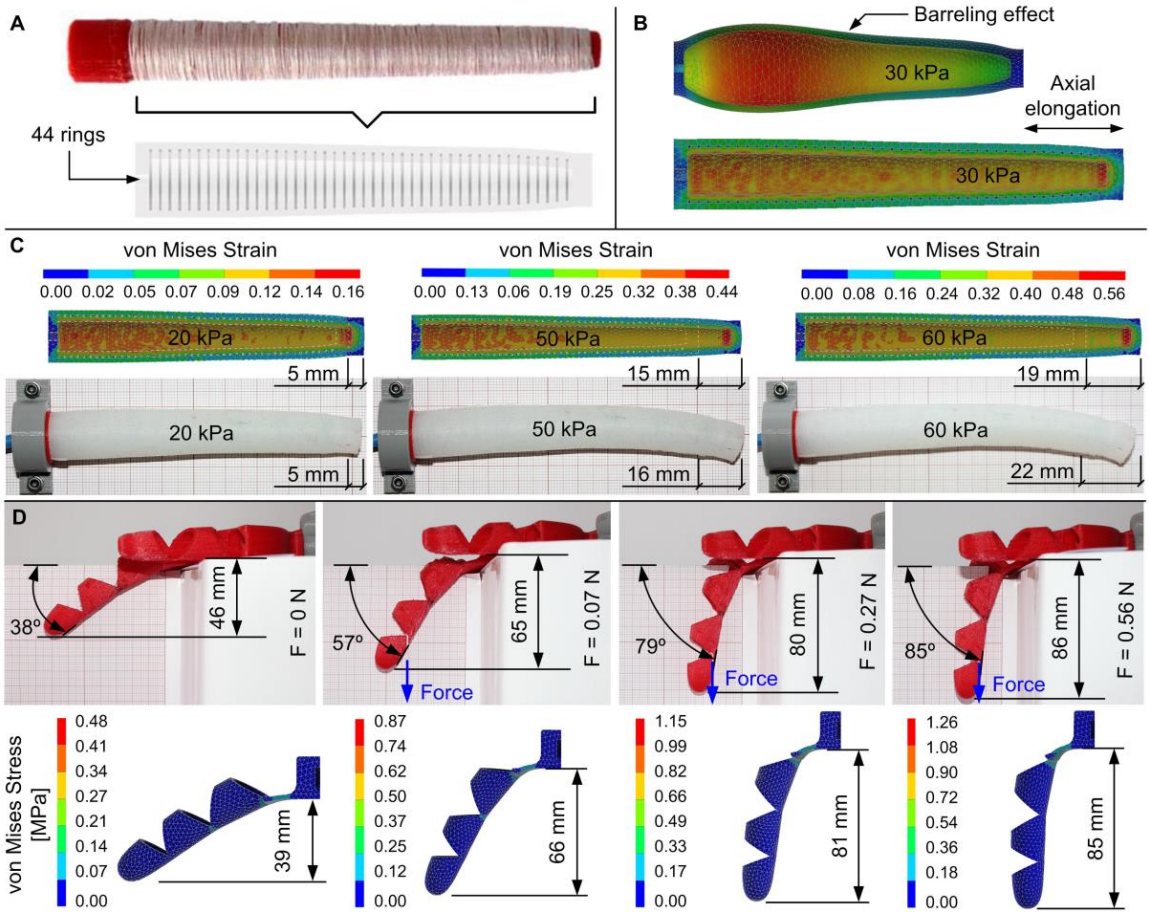

**Fig. 2**. Numerical modeling and simulation of soft actuator and exoskeleton. (A) The PET reinforcement around the mold's core in the top image, while the bottom image shows a simplified model with a cylindrical cross-section measuring 0.25 mm in diameter and consisting of 44 rings. (B) Effect of the PET reinforcement on the elimination of the barreling effect. (C) Comparison between experimental and numerical (Ogden set3) elongation of the actuator for three values of applied pressure. (D) Comparison between experimental and numerical (MR set3) bending of the exoskeleton finger subject to the gravity effect and force applied at the distal joint. The maximum stress value at the proximal joint falls within the material limits. When there is no external force applied, the exoskeleton weight causes vertical displacement.

## 2.2 Fabrication

The actuators were fabricated using molding while the exoskeleton was 3D printed using Fused Filament Fabrication (FFF), Fig. 3A. The 3D printing of the exoskeleton is a single-step process that facilitates and speeds up the fabrication of the most complex element of the hand, the exoskeleton, Fig. 3B. The five actuators and their corresponding air tubes are assembled inside the exoskeleton (Movie S2, Supplementary Materials). The silicone in a liquid state is poured into the mold where the PET reinforcement is winding the mold core, followed by a period of about 2 minutes in the vacuum chamber to eliminate air bubbles. After 10 hours at room temperature, the elastomer gets solid and is removed from the mold, as well as from the mold core. To add the internal elastomer layer, the unfinished actuator is placed again in another mold, vacuum chamber, and after the silicone solidifies, it is removed from the mold. A PLA-made rigid ring is glued to the base of the soft actuator and an air tube is connected to the ring. The assembly process consists of inserting the five actuators inside each exoskeleton finger, placing each ring at the ring holder of the exoskeleton, Fig. 3B. Finally, the five air tubes are connected to the valves.

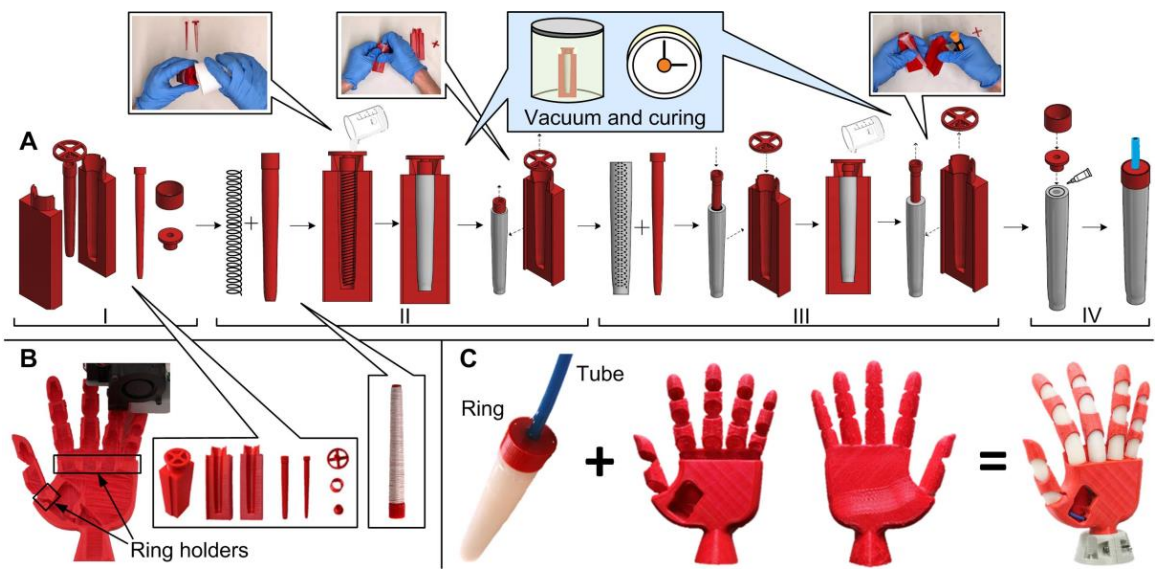

**Fig. 3**. Fabrication of the soft actuators and exoskeleton. (A) Actuator's fabrication steps using molding. In phase I, the molds and their components are 3D printed in PLA. In phase II, the PET reinforcement is wound around the mold core, the liquid state silicone is poured into the mold, and after a period in the vacuum chamber and curing at room temperature, the unfinished actuator is removed from the mold. In phase III, the unfinished actuator is placed again in another mold to add the internal silicone layer, and after another period in the vacuum chamber and curing at room temperature, the actuator is removed from the mold. In phase IV, the actuator is glued to a rigid ring to connect the air tube and fix the actuator to the exoskeleton ring holders. (B) The exoskeleton is 3D printed in a single-processing step using FFF. (C) The five actuators are inserted inside each exoskeleton finger with the ring placed on the exoskeleton holder.

## 2.3 Control

Finger's motion is controlled using low-cost off-the-shelf hardware components, aiming to promote scalability and reproducibility, Fig. 4A. The control board (CB) includes a microcontroller unit that runs a closed loop ON-OFF controller that commands the solenoid valves to inflate or deflate the actuators according to the set bending angles, Fig. 4B. The CB and related electronics control the inner pressures related to the set bending angles estimated in the FEA. Each one of the three controllable elements (thumb finger, index finger, and middle-ring-little fingers) is composed of 2 valves and a monitoring pressure sensor.

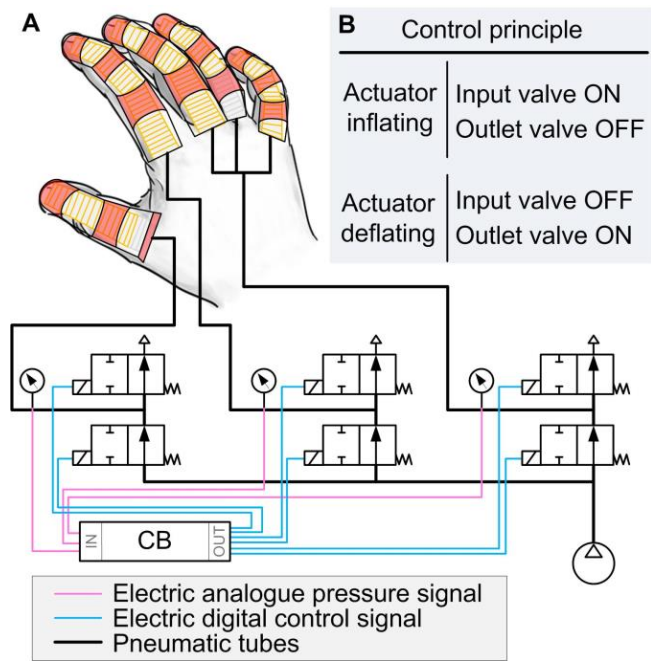

**Fig. 4**. Control/monitoring hardware and ON-OFF control principle. (A) Hardware components of the control system with the control commands labeled in blue and the monitoring pressure signals labeled in red. (B) The ON-OFF control principle for the three independent actuators, each one controlled by two valves and monitored by a pressure sensor.

3. **Experiments and Analysis**

3.1 **Materials**
   The actuator is made of platinum-catalyzed silicone (Ecoflex 00-50, Smooth-On, USA) reinforced with threads of polyethylene terephthalate (PET) yarn. The exoskeleton is made of flexible thermoplastic polyurethane (TPU) material (NinjaFlex, NinjaTek, USA). The exoskeleton was 3D-printed using a FFF 3-axis single-nozzle machine (Prusa i3 MK3S+, Prusa, Czechia). The printer G-code is generated from a slicer (PrusaSlicer 2.5.0, Prusa, Czechia), having the computer-aided design (CAD) model of the exoskeleton as input. The part's models were developed in CAD (Inventor 2021, Autodesk, USA). The molds to fabricate the actuator were 3D printed in the same machine using the same slicing software. The molds are made of polylactic acid (PLA) material (polylactic acid, Prusa, Czechia).

3.2 **Finite element analysis of hyper-elastic models**
   A commercial finite element package was used in the numerical analysis (Inventor Nastran, Autodesk, USA). Three different simulations were considered: (i) analysis of the actuator; (ii) analysis of the exoskeleton; (iii) analysis of the complete finger (actuator and exoskeleton). Only half geometry of a sample finger was modeled taking advantage of the symmetric conditions, Fig. S1, Supplementary Materials, allowing to reduce the computation cost. Both the exoskeleton and the actuator were modeled with quadratic tetrahedral finite elements (solid elements with 10 nodes). The mechanical behavior of all materials was described by hyper-elastic models.

The pressure was applied incrementally at the inner surface of the actuator while the base was fixed. The PET reinforcement was modeled considering 44 rings equally spaced along the longitudinal direction. To simplify the mesh generation stage, the cross-section of the rings was considered rectangular with 0.25 mm². Since each ring presents 4 contact surfaces, 176 contact interfaces were established. The contact between the silicone and the reinforcement rings was defined as bonded to prevent separation or relative movement between them. The average element size used in the mesh of the reinforcement was 1 mm while for silicone it was 1.5 mm. The mesh is presented in Fig. S1A. Since the stiffness of the PET reinforcement is higher than the silicon stiffness, a linear elastic behavior was assumed for the reinforcement material, using a mass density $\rho$ = 1541 kg/m³, elastic modulus E = 2.76 GPa, Poisson's ratio $\nu$ = 0.417 and the yield stress $\sigma_y$ = 5.44 GPa. The material of the actuator was defined by the Ogden model, considering three different sets of parameters, Table S1. Since the stiffness is affected by the air bubbles that may exist inside the material, the parameters related to the material stiffness were adjusted. The Ogden set1 parameters are the ones available in the literature [49] while the Ogden set2 and set3 parameters were adapted according to the properties of the material after fabrication.

The exoskeleton of a sample finger was fixed at the end, near the proximal joint, and a vertical load was applied at the distal joint Figure S1B, aiming to assess the bending stiffness of the exoskeleton. Accordingly, the vertical displacement was measured at the tip of the finger for different values of applied force. The average element size used in the mesh of the exoskeleton (a single finger) was 2.5 mm. The mesh is presented in Fig. S1B. The material of the exoskeleton (thermoplastic polyurethane) was defined by the MR model, considering three different sets of parameters, Table S2. After 3D printing, the material presents smaller density values, so we considered 0.83 g/cm³ for both MR set2 and MR set3.

### 3.3 Frictional contact

The analysis of the complete finger (exoskeleton and actuator) requires the definition of the frictional contact between the surfaces. The master-slave approach (unsymmetric contact) was adopted in the discretization of the contact interface, which requires less computational effort. Since the mesh size of the exoskeleton is larger than the mesh used for the actuator, the master surface was assigned to the exoskeleton surface, while the external surface of the actuator is defined as the slave surface. The master surface was divided into 10 regions, Fig. S1, aiming to define different contact conditions according to the contact pair interactions. The application of pressure in the chamber of the actuator yields mainly longitudinal deformation and consequently promotes some sliding between the actuator and the exoskeleton. Hence, the occurrence of sliding without separation was assigned to the regions identified by 1, 4, 6, 8, 9 and 10 in Fig. S1D. Nevertheless, the bending of the finger causes the separation of some regions of the exoskeleton in relation to the actuator. Accordingly, the regions identified by 2, 3, 5, and 7 were modeled as separation contact. During the incremental application of the internal pressure in the actuator, the end near the proximal joint was fixed. Regarding the hyper-elastic behavior of both actuator and exoskeleton materials, the set of constitutive parameters used in the numerical analysis was obtained from the comparison between numerical and experimental results when the exoskeleton and the actuator were studied separately. Fig. S1E compares the numerical and experimental deformed configuration (bending) of the complete finger for different values of pressure, highlighting the strain values inside the actuator chamber.

### 3.4 Experimental mechanical tests

The experimental mechanical tests were conducted in similar environmental conditions using sample exoskeletons, actuators, and complete fingers. Each exoskeleton was evaluated on a trial of 3 tests for each load. Each actuator was evaluated on a trial of 3 tests for each inner pressure, while each complete finger was evaluated 3 times for each different inner pressure considered. The ground truth vertical displacements and the related bending angles were estimated from the analysis of the static image frames recorded by a camera (EOS 1300D 18-55IS, Canon, Japan). Millimeter paper was placed on the background to facilitate the readings. The exoskeleton was fixed horizontally on the finger's base to guarantee the same initial reference in all circumstances. The exoskeleton's vertical displacement was evaluated by applying loads of 0.07 N, 0.16 N, 0.27 N, 0.40 N, and 0.56 N. These loads were generated by attaching weights to the distal joint. The exoskeleton's weight and the gravity effect compose the effective efforts actuating on it. The actuators were fixed on the base and their horizontal displacement was evaluated by applying different inner pressures, from 20 kPa to 60 kPa. The complete fingers were fixed on their base and the vertical displacement was evaluated by applying different inner pressures, from 20 kPa to 50 kPa.

### 3.5 Experimental control tests

All the tests to evaluate the ON-OFF controller were conducted in similar environmental conditions. Each test was repeated 5 times on 2 identical fingers, with an interval time of 45 minutes to dissipate residual strain energy. Off-the-shelf accessible hardware components compose the setup. The compressed air delivered to the actuator's chambers was supplied by an off-board portable air compressor equipped with a pressure regulator (TE-AC 270/50/10, Einhell, Germany). The valves, sensors and microcontroller are powered by a programmable DC power supply (72-13360, TENMA, China). The valves are one-way two-position (ON/OFF) 6V mini solenoid valves (CY05820D, cydfx, China). We used 6 valves, 2 to each one of the 3 controllable elements. Each one of the 3 elements has an absolute pressure sensor (MPX4250AP, NXP Semiconductors, Netherlands) providing measurements at a sampling rate of 50 Hz. The control board (Nano, Arduino, Italy) runs the ON-OFF controller receiving data from the sensors and actuating the valves. Air tubes with 3 mm diameter were used to connect the elements. The ground truth angles were measured by a magnetic tracker sensor (Liberty, Polhemus, USA) attached to the hand's finger. Data analysis was performed in MATLAB (MATLAB 2019b, MathWorks, USA).

## 4. Results

### 4.1 Mechanical behavior

The comparison between the numerical and experimental elongation of the actuator as a function of the internal pressure is presented in Fig. 5A. Using the Ogden set3, the numerical predictions are in good agreement with the experimental measurements. The variations observed at high-pressure levels could be attributed to the slight bending of the actuator observed in the experiments. The predicted von Mises strain distribution, presented in Fig. 2C, for three different values of applied internal pressure, demonstrated that the largest values arise at the inner surface of the actuator due to the large stiffness of the PET reinforcement in comparison with the material of the actuator. Indeed, the material of the actuator between the inner surface and the reinforcement rings is significantly compressed in the radial direction. Besides, the strain increases as the internal pressure increases, which promotes the elongation of the actuator and reduces the wall thickness. Regarding the exoskeleton, the numerical predictions are in good agreement with the experimental measurements of vertical displacement at the fingertip, Fig. 5B. The load-induced bending of the finger is primarily caused by the deformation occurring at the proximal joint, where the bending moment is high. As a result, the joints are where the exoskeleton of the finger experiences high-stress levels.

The FEA of the complete finger was carried out using the set of constitutive parameters selected in the previous FEA (actuator and exoskeleton) that provided the best accuracy. The numerical modeling of the frictional contact between the exoskeleton and the actuator is challenging to estimate due to the large sliding occurring between the bodies during the application of the internal pressure on the actuator. Therefore, the definition of the contact interface was adjusted to avoid convergence problems. The comparison between the numerical prediction and the experimental configuration of the complete finger is presented in Fig. 5C-D for three different input pressure values. The accuracy of the numerical model is evaluated through the vertical displacement at the fingertip. The bending of the complete finger is underestimated by the numerical simulation. This can be the consequence of improper modeling of the frictional contact between the exoskeleton and the actuator. Additionally, there are other sources of error, namely simplified boundary conditions, defects in fabrication, material memory, and thickness inhomogeneity, among other factors.

The force exerted by a sample finger on a rigid surface was evaluated, Fig. 5E. The finger is fixed at the base and installed at a distance of 40 mm to the force sensor. The force at the fingertip was measured for different pressures, showing a quasi-linear behavior that stabilizes at 80 kPa.

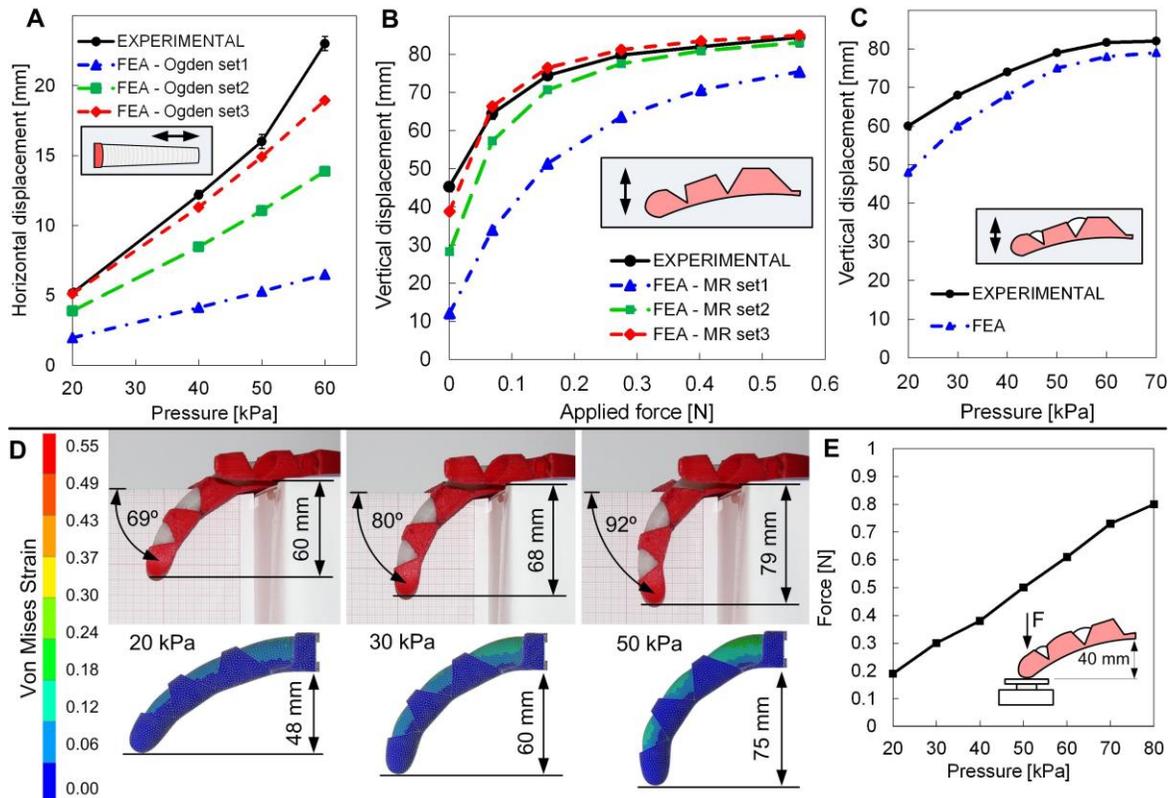

**Fig. 5**. Actuator, exoskeleton and complete finger numerical and experimental results. (A) Comparison between the numerical and experimental axial elongation of the actuator for different values of applied pressure. The simulation results present similar behavior to the experimental ones for the Ogden set3 model considering input pressures lower than 50 kPa. (B) Comparison between numerical and experimental deflection of the exoskeleton, evaluated for different values of force applied at the distal joint. (C) Comparison between numerical and experimental deflection of the complete finger for different values of applied pressure. (D) Comparison between numerical and experimental deformed configuration (bending) of the complete finger for different values of pressure. Strain values are within the limits of the materials. (E) Force measured at the fingertip for different pressure values.

## 4.2 Control

Experimental tests evaluated the actuation response of a sample index finger to different input pressures, angular range, ability to keep a set bending angle, and hysteresis. The actuator was pressurized with incremental pressures of 15 kPa for 5 seconds, from 0 to 45 kPa, and then the pressure was reduced, step by step following similar behavior, Fig. 6A. During the ascending steps, the pressure tends to decrease slightly under the set pressure, while during the descent steps the pressure tends to slightly rise. This phenomenon can be explained either by the hydraulic shock effect and the material stress relaxation. Overall, the system demonstrated the capability to reach and maintain the set pressures. Since the ON-OFF control demands high-frequency open/close actions from the valves to admit and expel air, we evaluated the ability of the control system to keep a set pressure stable for longer times, successfully maintaining the set bending angle, Fig. 6B. The pressure levels are marginally below the set values, as a consequence of having the input valves shut off as soon as the sensor reads those levels, creating a small pressure offset.

The hysteresis test consisted in bending the finger until it reached 75 kPa, resulting in a 95° bending angle, holding the pressure steady for 10 seconds, and then, decreasing the pressure until the actuator returned to the neutral pose, Fig. 6C. The angle was measured using a magnetic tracker sensor attached to the fingertip, Fig. 6E. Owing to the elastic hysteresis, the same inner pressure results in two different angles, depending on if the finger is opening (deflating) or closing (inflating). The perturbation, while the finger is opening, is a consequence of the non-uniform slipping that occurs on the actuator-exoskeleton contact surfaces close to the proximal joint.

The finger can maintain the set pressure even in the presence of air leaks in the actuator, as long as the input airflow is greater than the leakage flow. This is an interesting feature of the controller, making it possible to tolerate and recover from moderate air leakage. Fig. 6D shows the pressure values of an actuator with a leak, which after being inflated at 50 kPa, the air escapes rapidly until the atmospheric pressure is reached. When the ON-OFF controller is used, it can maintain the set pressure of 50 kPa with moderate oscillations. Every time the pressure drops, the inlet valve is activated, maintaining the desired pressure (Movie S3, Supplementary Materials).

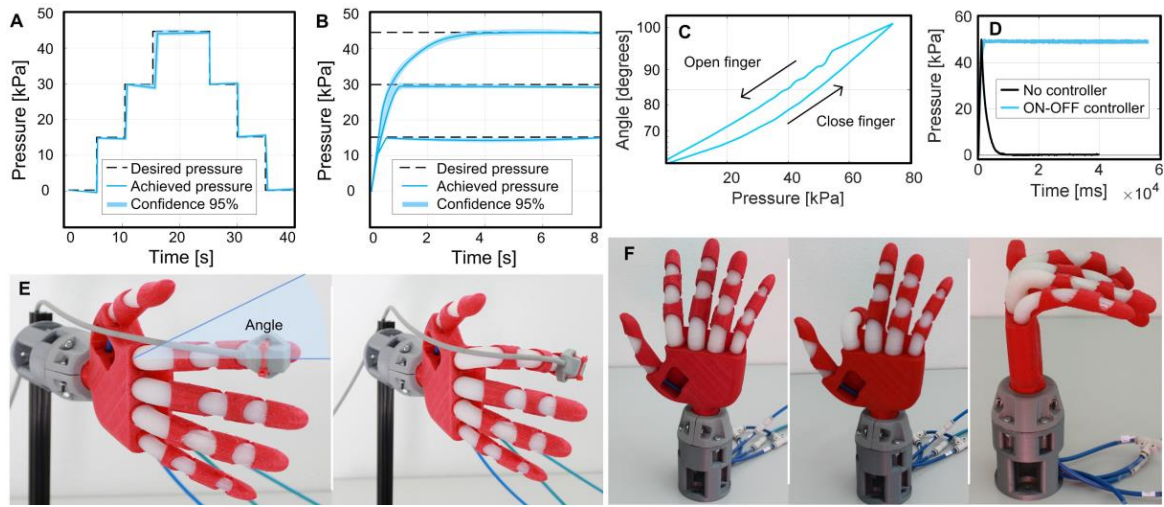

**Fig. 6**. Actuation response of a sample index finger. (A) System's response to step commands. Internal pressures are maintained within the different steps. (B) The control system allows keeping the set pressures for long periods. (C) Hysteresis results show a different behavior while closing and opening the finger. (D) The controller maintains the set pressure even in the presence of air leaks. (E) The magnetic tracker sensor is attached to the index finger to measure its angle (ground truth). (F) The robotic hand in a neutral pose, bending the index finger and with all fingers bent. In all graphs, the shaded area represents the standard deviation (95% confidence).

## 5. Conclusion and Discussion

This study presented the development, fabrication and control of a bio-inspired soft robotic hand. The FEA comprising (i) hyper-elastic behavior of the soft materials, (ii) finite rotation and large strain of the exoskeleton and actuators, and (iii) frictional contact between exoskeleton and actuators, demonstrated a valuable tool to support the design and control of the hand's fingers. The numerical predictions were in good agreement with the experiments, namely the relationship between the applied pressure and the deformed configuration of the fingers. As indicated in numerical results, the actuator's reinforcement in a circumferential direction, proven in experimental tests to guarantee the actuator's elongation and consequently the fingers bent when inside the exoskeleton. The robotic hand achieved an interesting dexterity level, being able to grasp objects of different shapes and sizes. Nevertheless, it struggles to grasp heavier objects featuring slippery surfaces, showing a concentrated deformation at the base of the fingers while the thumb motion is constrained. In addition, depending on the grasping surface and geometry, there exists mechanical interference between the fingers. Since the soft hand is highly nonlinear, with most variables of interest being coupled between themselves, future work will be dedicated to an in-depth analysis of the grasping phenomena together with further standardization of testing benchmarks. The ON-OFF controller guarantees the fingers are accurately bent to set angles and maintains the configuration for as long as necessary, even in the presence of instability (air leaks). This soft robotic hand is accessible and can be built at a reduced cost, avoiding the time-consuming design-fabrication trial-and-error processes, and inspiring innovation around it. The cost of the materials to fabricate the hand itself is around 6 dollars, plus the control elements (valves, pressure sensors, tubes and control board) which cost about 75 dollars. The equipment needed to fabricate it is a regular FFF 3D-printer and a vacuum chamber. Since the molds and the exoskeleton are 3D-printed, the fabrication of the complete hand takes about 14 hours, including the materials curing.

## Acknowledgments

**Author contributions:** S.A. and M.B. contributed to the design conceptualization, fabrication, and conducted the experimental tests. A.S. contributed to the control of the robotic hand. D.N. contributed to the numerical simulation. D.F. contributed to the analysis and result interpretation, and P.N. contributed to funding acquisition, review and writing.

**Funding:** This work was supported by Portuguese national funds through FCT - Fundação para a Ciência e a Tecnologia, [grant numbers UIDB/00285/2020, LA/P/0112/2020 and 2022.13512.BD].

**Competing interests:** The authors declare that there is no conflict of interest regarding the publication of this article.

**Data Availability:** The data that support the findings of this study are available from the corresponding author upon reasonable request.

## Supplementary Materials

Figure S1
Tables S1 to S2
Movies: https://spj.science.org/doi/full/10.34133/cbsystems.0051

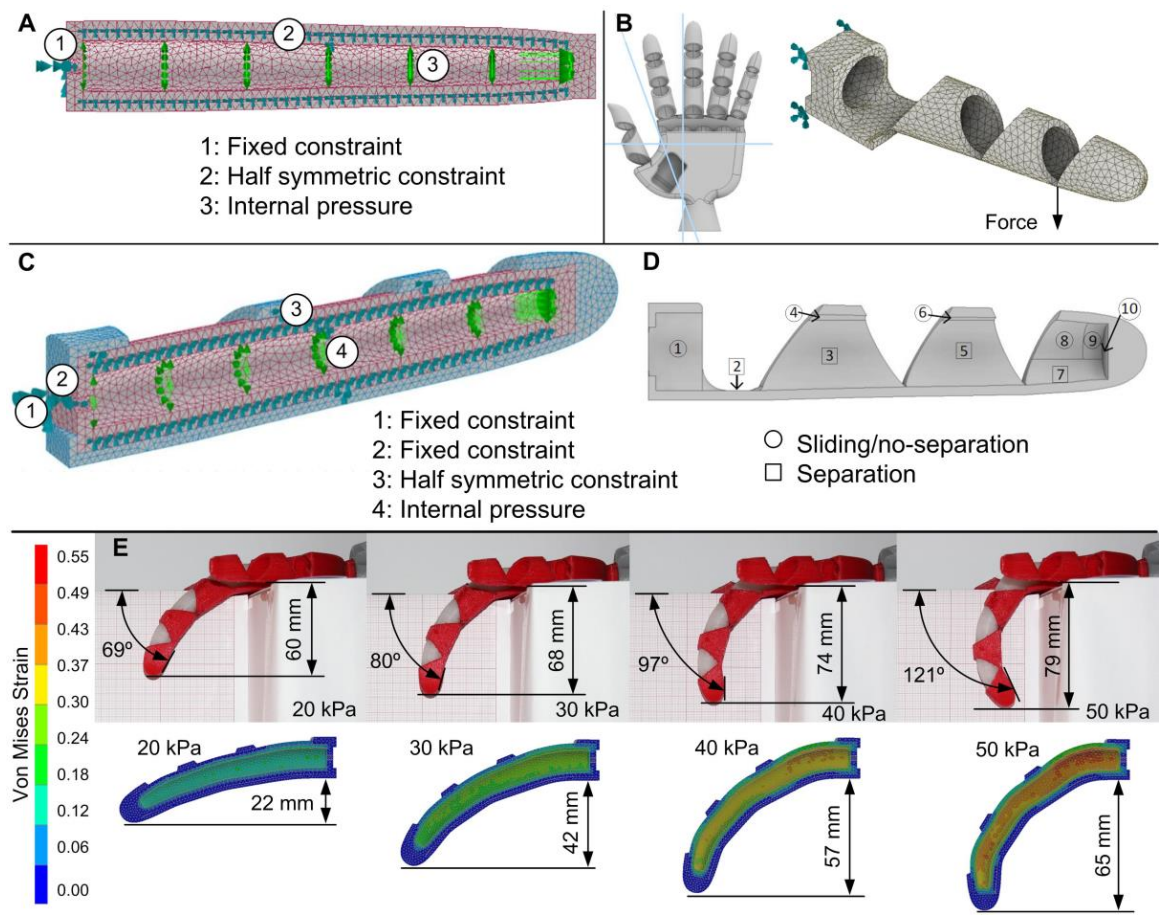

**Fig. S1**. FEA finger's constraints. (A) Applied constraints to the actuator, which include a fixed constraint on the base, a half-symmetric constraint, and the internal pressure on the chamber walls. (B) Exoskeleton FEA mesh. Each exoskeleton finger is fixed on the base and the load force applied at the distal joint. (C) Applied constraints to the complete finger. (D) Sliding/separation areas between the actuator and the exoskeleton. (E) Numerical and experimental configuration of the complete finger for different values of pressure, highlighting the strain values inside the actuator chamber.

**Table S1**. Three different sets of material parameters of the Ogden hyper-elastic model are used to describe the platinum-catalyzed silicone.

| Set | Parameters | | Incompressible value |
|---|---|---|---|
| Ogden set1 | $\alpha_1=1.55$ | $\mu_1=107900.00$ | $D_1=1\times10^5$ |
| | $\alpha_2=7.86$ | $\mu_2=21.47$ | |
| | $\alpha_3=-1.91$ | $\mu_3=-87100.00$ | |
| Ogden set2 | $\alpha_1=1.05$ | $\mu_1=1.50\times10^5$ | $D_1=1\times10^5$ |
| | $\alpha_2=4.00$ | $\mu_2=60.00$ | |
| | $\alpha_3=-1.60$ | $\mu_3=-1300.00$ | |
| Ogden set3 | $\alpha_1=1.05$ | $\mu_1=1.12\times10^5$ | $D_1=1\times10^5$ |
| | $\alpha_2=4.00$ | $\mu_2=45.00$ | |
| | $\alpha_3=-1.60$ | $\mu_3=-975.00$ | |

**Table S2**. Three different sets of material parameters of the MR hyper-elastic model are used to describe thermoplastic polyurethane.

| Set | Parameters [MPa] | | Material density [g/cm³] |
|---|---|---|---|
| MR set1 | $C_{10}=0.677$ | $C_{01}=1.621$ | 1.19 |
| MR set2 | $C_{10}=0.300$ | $C_{01}=0.750$ | 0.83 |
| MR set3 | $C_{10}=0.210$ | $C_{01}=0.525$ | 0.83 |

**Movie S1**. Object Grasping and Manipulation

The soft robotic hand is attached to a robot manipulator (MOTOMAN SIA, Yaskawa, Japan), grasping and manipulating different objects from the Yale-CMU-Berkeley (YCB) dataset. It successfully grasps and manipulates the objects (fruits models and plastic/ceramic cups) at a robot speed of 100 mm/s with no visible slippage.

**Movie S2**. Fabrication

The fabrication of the soft robotic hand follows a sequential process, starting from the CAD models of the exoskeleton and actuators to their fabrication. The exoskeleton is 3D printed in a single part using FFF. The fabrication of the actuators using molding follows a sequential process. Molds and related components are 3D printed in PLA using FFF. The PET reinforcement is wound around the mold core, and the liquid-state silicone is poured into the mold. After ~2 minutes in the vacuum chamber and 10 hours of curing at room temperature, the unfinished actuator is removed from the mold and the mold core. This process is repeated in a second molding stage to add the internal elastomer layer.

**Movie S3**. Control

The hand fingers, controlled by the proposed ON-OFF control principle, keep the set bending angles for the index finger alone and the middle-ring-little fingers together. When the hand is not pneumatically actuated, it stays in a neutral pose. The ON-OFF controller regulates the airflow to the actuators, activating them and enabling smooth, human-like motion.